%% file: main.tex
\documentclass[10pt,twocolumn,letterpaper]{article}

\PassOptionsToPackage{sectionbib}{natbib}
\usepackage[pagenumbers]{cvpr}
\usepackage{chapterbib}
\usepackage{placeins}

\definecolor{cvprblue}{rgb}{0.21,0.49,0.74}
\usepackage[pagebackref,breaklinks,colorlinks,allcolors=cvprblue]{hyperref}
\newcommand{\suppref}[1]{\ref*{#1}}
\def\confName{CVPR}
\def\confYear{2027}

\title{When Extreme Darkness Meets Motion Blur: MeanFlow for Unified RAW Restoration}
\input{authors}

\makeatletter
\patchcmd{\@maketitle}{\vskip .375in}{\vskip 0pt}{}{\PackageError{main}{Could not adjust the preprint title spacing}{}}
\patchcmd{\@maketitle}{\vspace*{24pt}}{\vspace*{6pt}}{}{\PackageError{main}{Could not adjust the preprint author spacing}{}}
\makeatother

\begin{document}
\include{paper}
\include{supplementary}
\end{document}

%% file: authors.tex
\author{Zepu Wang\thanks{Equal contribution.}\quad
Jingze Liang\footnotemark[1]\quad
Weijie Xiao\footnotemark[1]\quad
Kexin Chen\\
University of Electronic Science and Technology of China (UESTC)
}

%% file: paper.tex
\maketitle
\input{sec/0_abstract}
\input{sec/1_intro}
\input{sec/2_related_work}
\input{sec/3_dataset}
\input{sec/4_method}
\input{sec/5_experiments}
\input{sec/6_conclusion}

\section*{LLM Usage Statement}
We used OpenAI Codex and Anthropic Claude to help debug research code. During writing, they suggested phrasing and grammar edits, reviewed argument flow, and flagged ambiguous statements or overstated claims. We drafted the manuscript ourselves and used LLM-generated wording only for brief, localized revisions. Throughout the process, we evaluated the suggestions critically and made our own decisions about the manuscript's content and argument.

The idea for this study came directly from our firsthand observations while using low-light enhancement models. We conceived the study and formulated the research question ourselves, independently of any AI input. When we applied these models to motion-blurred inputs in extreme darkness, we found that they could improve exposure while leaving edges and textures smeared. This observation led us to ask whether a single RAW restoration model could enhance sharp low-light inputs and jointly enhance and deblur motion-degraded inputs. Once the project was underway, we used LLMs to help locate relevant papers and websites. We spent substantial time investigating the question by constructing the paired RAW benchmark, developing the restoration model, and evaluating it across illumination levels. We designed and ran the reported experiments and take responsibility for the accuracy of the results and conclusions.

{\small
\bibliographystyle{ieeenat_fullname}
\bibliography{main}
}

%% file: sec/0_abstract.tex
\begin{abstract}

Extremely low-light enhancement aims to restore severely underexposed images to clean, well-exposed counterparts. However, existing methods predominantly focus on illumination loss and sensor noise, overlooking the camera shake that frequently accompanies handheld capture under extreme darkness. To address this problem, we propose LEAD-MF, an effective and efficient framework that enhances images across varying extremely low-light levels while handling the accompanying motion blur, and switches between clean enhancement and joint enhancement--deblurring within a single model. Operating in the RAW domain, LEAD-MF employs a tokenizer equipped with domain--illumination joint affine modulation (DIJAM) to map RAW observations from different exposure domains and darkness levels into a shared compact latent space, enabling efficient full-image inference. In this latent space, a unified conditional MeanFlow backbone predicts dual enhancement and deblurring velocity residuals, brightening the image while removing potential motion blur in a single step. We further construct SIDED, a paired RAW benchmark synthesized with realistic motion blur, in which each scene provides a sharp well-exposed reference, its blur-free low-light observation, and motion-blurred low-light and well-exposed counterparts. By integrating motion blur before sampling calibrated sensor noise, SIDED preserves the statistics of the calibrated RAW noise model. Extensive experiments from 0.01 to 0.0001 lux show that LEAD-MF outperforms existing enhancement-only, joint, and cascaded solutions on both SIED and SIDED at every illumination level, exceeding the strongest baseline by up to 2.68 dB PSNR.

\end{abstract}

%% file: sec/1_intro.tex
\section{Introduction}

Extremely low-light image enhancement aims to recover clean, well-exposed images from photographs captured under severely insufficient illumination, where only a few photons reach the sensor and the faint scene signal is buried in noise. Such enhancement has conventionally been performed on rendered RGB images, where Retinex decomposition, learned curves, and restoration networks improve brightness, contrast, and color \cite{Wei_2018_RetinexNet,Zhang_2019_KinD,Guo_2020_CVPR}. However, by the time an RGB image is produced, the camera ISP has already transformed the sensor measurements through demosaicing, denoising, white balance, color correction, tone mapping, and quantization. Under extreme darkness, these operations suppress weak sensor responses and entangle noise with rendering artifacts, discarding information that an RGB-domain model cannot recover. RAW-domain enhancement has therefore attracted increasing attention, since linear, high-bit-depth sensor measurements retain faint signals and camera-dependent noise statistics. SID first established paired restoration from short-exposure RAW images \cite{Chen_2018_CVPR}, and subsequent work has improved noise modeling \cite{Wei_2020_CVPR}, efficiency \cite{lamba2020fastlightweightrestorationdark,Lamba_2021_CVPR}, and network design \cite{Jin_2023_CVPR,Chen_2026}. More recently, SIED extended paired RAW enhancement to even darker illumination levels down to $0.0001$ lux \cite{Jiang_2025_ICCV}.

\begin{figure*}[!t]
    \centering
    \includegraphics[width=\textwidth,trim=0bp 215bp 0bp 0bp,clip]{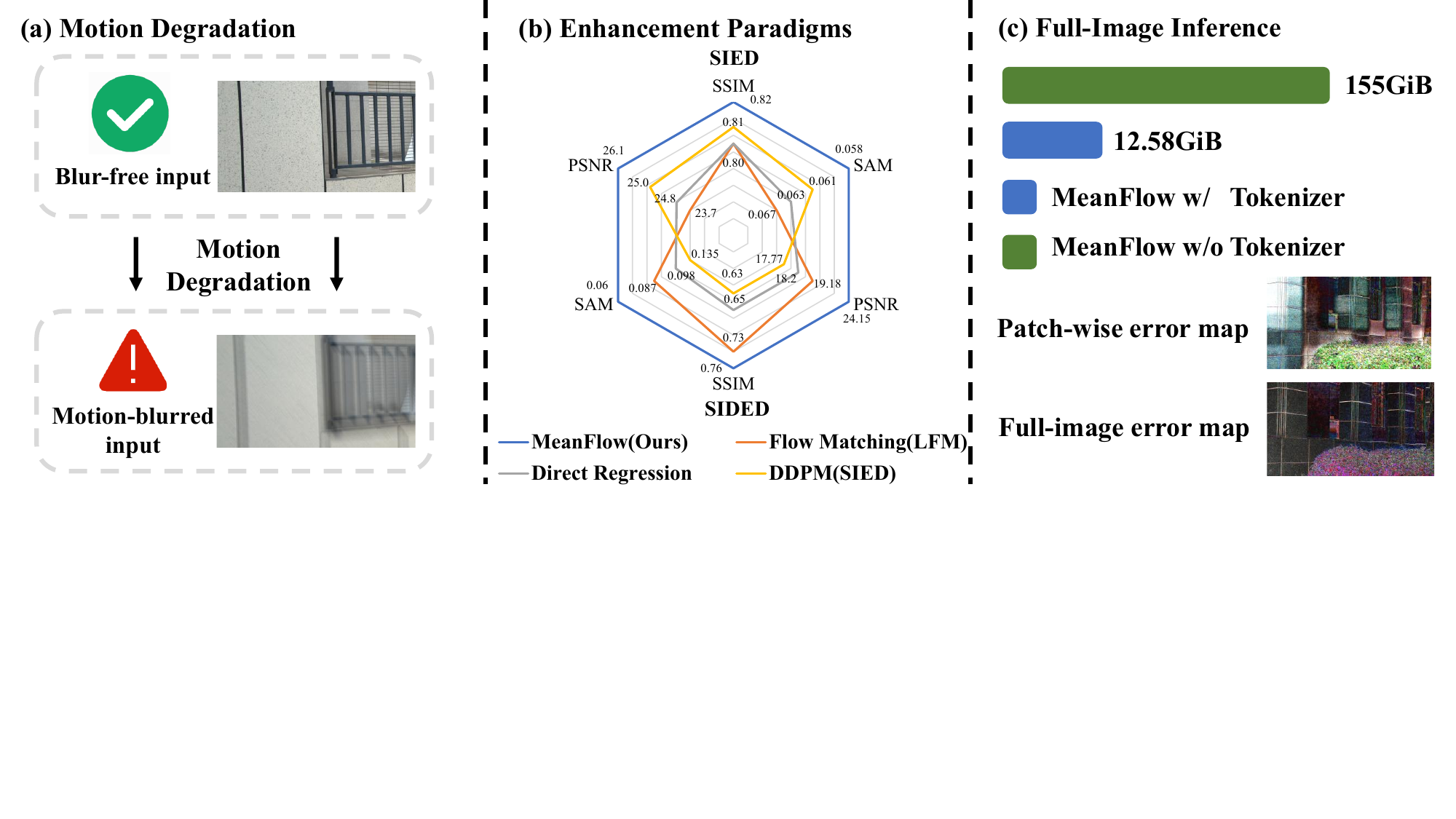}
    \caption{Motivation for unified full-image RAW restoration. (a) Results of an enhancement-only model that ignores camera motion, applied to a blur-free and a motion-blurred low-light input of the same scene. Although both outputs are well exposed, the motion-blurred case remains blurry, with smeared edges and railing structures. (b) Radar comparison of restoration paradigms for $0.01$-lux clean enhancement on SIED. MeanFlow (ours) and direct regression use the same tokenizer, backbone, and curriculum, whereas flow matching (LFM) and DDPM (SIED) denote the LFM-LLIE and SIED baselines; higher PSNR and SSIM and lower SAM are better. (c) The bars compare the full-image memory use of MeanFlow with and without tokenization. The bottom-right images are amplified absolute RGB error maps between reconstruction and ground truth for patch-wise and full-image inference on a $0.01$-lux test scene.}
    \label{fig:intro-motivation}
\end{figure*}

Despite advances in exposure correction, denoising, and color restoration across both domains \cite{Chen_2018_CVPR,Jin_2023_CVPR,Chen_2026,Jiang_2025_ICCV}, most extremely low-light enhancement studies concentrate on photometric degradation and sensor noise, while the blur accompanying handheld capture is often overlooked. In practical extremely low-light photography, insufficient illumination is also frequently accompanied by unintended camera motion, which smears edges and textures while severe sensor noise further obscures them \cite{Zhou_2022_LEDNet,Zhao_2022_D2HNet}. As illustrated in Fig.~\ref{fig:intro-motivation}(a), an enhancement model that does not account for camera motion restores the exposure of a motion-blurred
input, yet the result remains blurry: brightening alone does not recover the structures smeared by camera motion \cite{Zhou_2022_LEDNet,Lv_2024_CVPR}. This gap limits the applicability of existing approaches to practical extreme-darkness scenarios, where illumination loss, noise, and motion degradation frequently coexist.

To address this practical yet underexplored problem, we introduce See in the Degraded Extremely Dark (SIDED) and Low-light Enhancement And Deblurring with MeanFlow (LEAD-MF). SIDED extends the paired SIED benchmark \cite{Jiang_2025_ICCV} with carefully designed motion-degraded RAW observations that follow realistic camera trajectories and preserve RAW-domain noise characteristics, while retaining the sharp, well-exposed references so that clean enhancement and joint enhancement--deblurring can be evaluated on corresponding scenes. Specifically, motion synthesis starts from SIED's illumination-aligned signal before its calibrated noise is added \cite{Jiang_2025_ICCV}. Following stochastic point-spread-function trajectories \cite{Boracchi_2012_TIP}, we simulate continuous, variable-speed camera paths that resemble handheld shake during long exposures in the dark. The calibrated noise is then sampled after motion integration, following the blur-then-noise ordering of RSBlur \cite{Rim_2022_ECCV}, so that the blur simulation does not filter the noise and alter its statistics. For this coupled setting, LEAD-MF pairs a shared RAW tokenizer with a task-conditioned, unified MeanFlow backbone to restore both blur-free and motion-degraded inputs within a single model.

To achieve this, LEAD-MF is built around two stages. First, we design a shared RAW tokenizer equipped with Domain--Illumination Joint Affine Modulation (DIJAM), which conditions on the acquisition domain and continuous log-illumination to map multi-level low-light and well-exposed RAW images into an aligned, compact latent space, as verified by the DIJAM ablation in Sec.~\ref{sec:abl-dijam}. This latent design preserves the full-image consistency to which extremely low-light enhancement is highly sensitive, avoiding the larger and spatially uneven errors of patch-wise inference while cutting full-image memory usage from $155$~GiB to $12.58$~GiB (Fig.~\ref{fig:intro-motivation}(c)), thereby making full-image MeanFlow feasible. Second, within this latent space, conditional MeanFlow \cite{NEURIPS2025_6d13e085} learns task-conditioned average velocities along paired restoration paths and predicts separate enhancement and deblurring residuals for one-step restoration within a unified backbone, as verified by the dual-head and deblurring-output ablations in Secs.~\ref{sec:abl-heads} and~\ref{sec:abl-deblur}. As shown in Fig.~\ref{fig:intro-motivation}(b), LEAD-MF outperforms direct regression under the same tokenizer and backbone, as well as the flow-matching LFM-LLIE and diffusion-based SIED baselines.

Experiments at three extreme illumination levels demonstrate the effectiveness of LEAD-MF. On both SIED and SIDED, it ranks first in PSNR, SSIM, and Spectral Angle Mapper (SAM) among the reported methods at every level. On SIDED, its PSNR exceeds the strongest reported baseline by $2.68$, $2.52$, and $2.24$ dB at $0.01$, $0.001$, and $0.0001$ lux, respectively.

The main contributions of this paper are summarized as follows:

\begin{itemize}
    \item We introduce SIDED, a realistic paired RAW benchmark that combines trajectory-based camera motion and calibrated extreme-low-light noise. It models realistic motion before the added-noise sampling step and retains sharp references for controlled evaluation.

    \item To the best of our knowledge, we are the first in extremely low-light RAW enhancement to explicitly address the motion blur that commonly accompanies extreme-darkness capture, establishing both a dedicated paired dataset and an effective, efficient unified framework (LEAD-MF) to tackle clean enhancement and joint enhancement--deblurring within a single model.

    \item We design a unified architecture that employs a DIJAM-equipped RAW tokenizer to map multi-level low-light and well-exposed RAW inputs into a shared compact latent space, coupled with a conditional latent MeanFlow backbone for one-step brightening and deblurring. Furthermore, we incorporate Retinex supervision during training to align the restored images with physical priors, achieving superior performance across all benchmarks and illumination levels.
\end{itemize}

\begin{figure*}[!t]
    \centering
    \includegraphics[width=\textwidth]{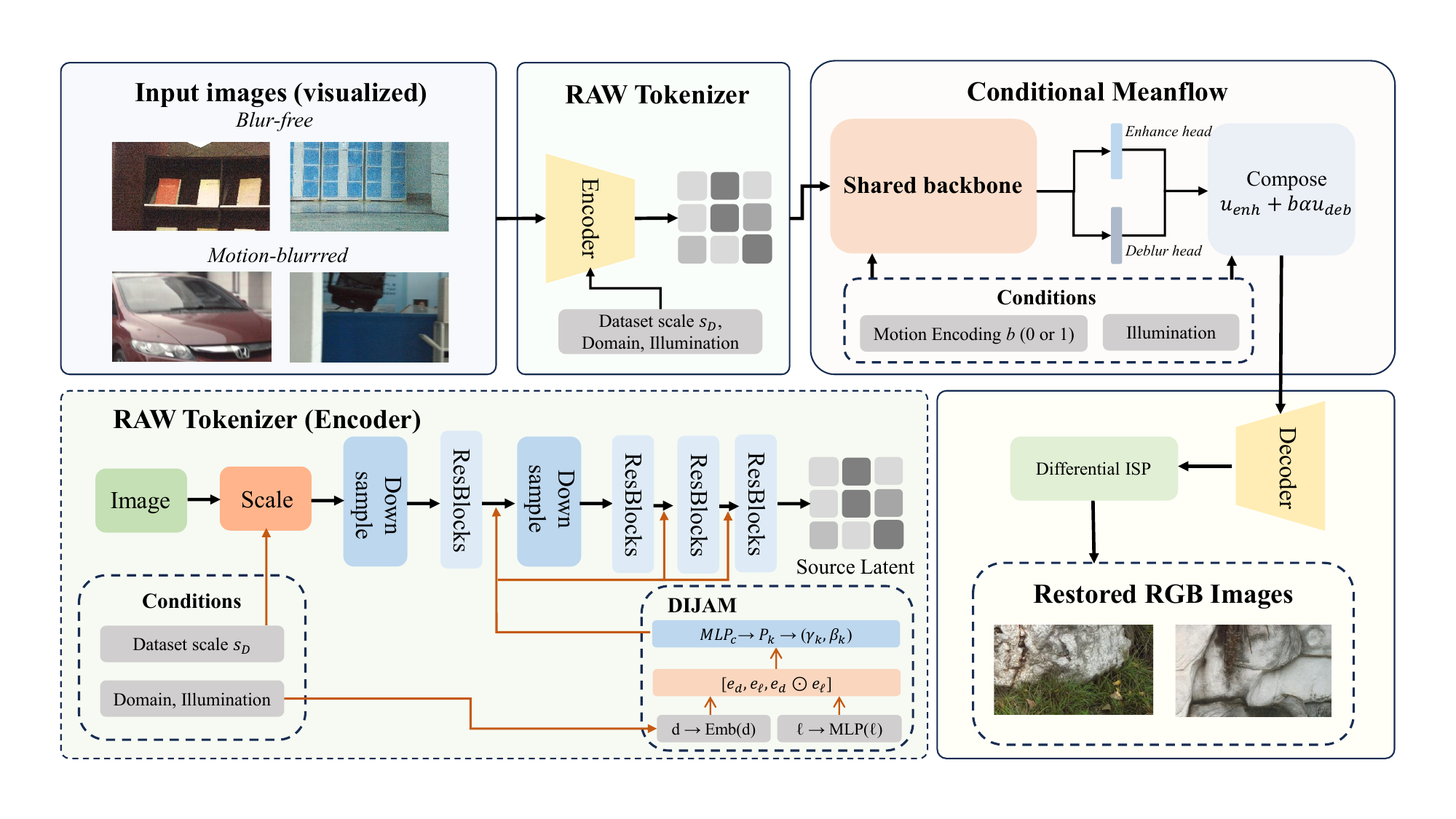}
    \caption{Overview of LEAD-MF at low-light inference. The RAW tokenizer encodes the input into a source latent, with domain--illumination conditioning applied across encoder stages. A unified conditional MeanFlow backbone predicts enhancement and deblurring residuals; the task condition determines their composition. For the clean-enhancement and joint-restoration paths shown, $e=0$ and $\alpha=b$ ($b=0$ and $b=1$, respectively), so $e$ and $\alpha$ are omitted from the diagram. The predicted latent is decoded to RAW and rendered to RGB by the frozen differentiable ISP. The lower-left panel expands the tokenizer encoder and its conditioning pathway.}
    \label{fig:framework}
\end{figure*}

%% file: sec/2_related_work.tex
\section{Related Work}

\paragraph{Extremely low-light RAW enhancement.}
Paired RAW enhancement recovers well-exposed images from short-exposure sensor measurements \cite{Chen_2018_CVPR}; subsequent work improves noise modeling, Bayer-domain representation, and efficiency \cite{Wei_2020_CVPR,lamba2020fastlightweightrestorationdark,Jin_2023_CVPR,Chen_2026}, and SIED extends evaluation to more severe darkness \cite{Jiang_2025_ICCV}. These works assume blur-free inputs, whereas we also address motion-degraded observations of the same scenes.

\paragraph{Joint low-light enhancement and deblurring.}
Deblurring has advanced from coarse-to-fine and transformer architectures to frequency-, motion-, and state-space-driven designs \cite{Cho_2021_ICCV,Zamir_2022_CVPR,Kong_2023_CVPR,Liu_2024_CVPR,Kong_2025_EVSSM}, while joint low-light enhancement and deblurring methods employ encoder--decoder decoupling, hierarchical restoration, Fourier-prior diffusion, and auxiliary structural tasks \cite{Zhou_2022_LEDNet,Zhao_2022_D2HNet,Lv_2024_CVPR,Li_2024_TMM,Feijoo_2025_DarkIR,Liu_2025_LIEDNet}. However, they target conventional low light rather than extreme darkness and operate on ISP-rendered RGB images whose weak signals are already suppressed and entangled with noise; under extremely low SNR compounded by motion blur, they thus struggle to recover faithful structures (Table~\ref{tab:sided-quantitative}), motivating our unified RAW-domain restoration.

\paragraph{Latent restoration.}
Diffusion restoration iteratively refines corrupted images or latents \cite{Yi_2023_ICCV,Kawar_2022_DDRM,Lin_2025_DiffBIR,Zhu_2026_RTD}, flow matching models distribution transport \cite{lipman2023flowmatchinggenerativemodeling} (\eg, LFM-LLIE for iterative RGB enhancement \cite{Ciubotariu_2026_CVPRW}), and MeanFlow enables one-step average-velocity prediction \cite{NEURIPS2025_6d13e085}. Output decomposition appears in RDDM (residual and noise) \cite{Liu_2024_RDDM}, iMF (an auxiliary velocity head) \cite{Geng_2026_iMF}, and FRSE (speech flow matching) \cite{Cai_2026_FRSE}. While encoder--decoder latents reduce restoration cost \cite{Rombach_2022_CVPR,Jiang_2025_AutoDIR,Cao_2026_TOP}, paired low-light and well-exposed RAW images exhibit a severe domain gap; inspired by feature modulation and domain normalization \cite{Perez_2018_FiLM,Chang_2019_CVPR,Liang_2026_ReTailor}, DIJAM jointly conditions on RAW domain and continuous illumination.

\paragraph{Physical priors and Retinex theory.}
Retinex theory decomposes an image into illumination and reflectance and has been adopted for low-light enhancement \cite{Wu_2022_CVPR,Cai_2023_ICCV}, diffusion restoration \cite{Yi_2025_TPAMI}, and underwater restoration \cite{Varghese_2025_CVPR}. Unlike methods that embed Retinex decomposition into inference, we employ Retinex consistency \cite{Yi_2025_TPAMI} only as a training-time prior in the final refinement phase, improving fidelity without inference overhead, as verified by the Retinex ablation in Sec.~\ref{sec:abl-retinex}.

%% file: sec/3_dataset.tex
\section{SIDED Dataset}

\paragraph{Source pairs.}
SIDED builds upon SIED \cite{Jiang_2025_ICCV}, which provides paired extremely low-light observations and sharp well-exposed references across three illumination levels. Starting from SIED's noise-free, illumination-aligned RAW signal $I_{\mathrm{syn}}$, we further subtract a dark-frame map $D_{\mathrm{ref}}$ measured with a sensor of the same model to obtain the clean intermediate $I_{\mathrm{syn}}^{\mathrm{df}}=I_{\mathrm{syn}}-D_{\mathrm{ref}}$, while retaining the original blur-free pair $(x_l, x_g)$ for clean enhancement evaluation.

\paragraph{Realistic camera motion.}
Rather than a fixed linear kernel with uniform velocity, we model handheld shake via stochastic point-spread-function trajectories \cite{Boracchi_2012_TIP}, \ie, a continuous sensor-plane path $p(t)$ over the exposure interval $[0,T]$. A smooth, positive speed profile reparameterizes exposure time, so the trajectories vary in displacement, orientation, curvature, and speed. For a scene signal $I$, the motion-integrated signal is
\begin{equation}
\mathcal{B}_{p}(I)=\frac{1}{T}\int_0^T \mathcal{W}(I;p(t))\,\mathrm{d}t,
\end{equation}
where $\mathcal{W}$ warps $I$ by $p(t)$, evaluated numerically over temporally sampled warps. Both low-light and well-exposed signals share the same path $p(t)$ to remain strictly paired. Sampled trajectories, their PSFs, and the resulting blur are visualized in Fig.~\suppref{fig:supp-motion-previews}, with full construction details in Sec.~\suppref{sec:supp-sided} of the supplementary material.

\paragraph{Post-motion noise sampling and dataset contents.}
Directly blurring a noisy RAW image averages out high-frequency fluctuations and introduces spatial correlations, distorting the sensor noise statistics. Following the blur-then-noise ordering of RSBlur \cite{Rim_2022_ECCV}, we sample calibrated sensor noise after motion integration:
\begin{equation}
x_l^{\mathrm m}=\mathcal{N}\!\left(\mathcal{B}_{p}(I_{\mathrm{syn}}^{\mathrm{df}});\kappa\right),
\end{equation}
where $\mathcal{N}(\cdot;\kappa)$ denotes SIED's calibrated physics-based noise sampler \cite{Jiang_2025_ICCV,Wei_2020_CVPR} under camera and ISO condition $\kappa$, combining signal-dependent shot noise conditioned on $\mathcal{B}_{p}(I_{\mathrm{syn}}^{\mathrm{df}})$, Gaussian read noise, and sampled dark-frame noise. Each scene in SIDED thus provides four aligned RAW observations: the blur-free pair $(x_l,x_g)$ and their motion-degraded counterparts $(x_l^{\mathrm m},x_g^{\mathrm m})$, where $x_g^{\mathrm m}=\mathcal{B}_{p}(x_g)$; $x_g$ serves as the sharp well-exposed reference.

%% file: sec/4_method.tex
\section{Methodology}

\subsection{Overview}

As illustrated in Fig.~\ref{fig:framework}, LEAD-MF comprises a frozen DIJAM RAW tokenizer and a single task-conditioned MeanFlow backbone. For an input RAW image $x_{\mathrm{in}}$, the condition vector is
\[
c=(d,e,\ell,b,\alpha),
\]
where $d$ denotes the RAW domain, $e\in\{0,1\}$ the source exposure (low-light or well-exposed source), $\ell$ the normalized illumination level, $b$ blur presence, and $\alpha$ the deblurring strength. The tokenizer takes $c_{\mathrm{tok}}=(d,\ell)$ to encode $x_{\mathrm{in}}$ into a source latent $z_{\mathrm{in}}$, while the MeanFlow backbone takes $c_{\mathrm{mf}}=(e,\ell,b,\alpha)$.

\subsection{Stage I: RAW Tokenizer with DIJAM}

Because extremely low-light and GT RAW images exhibit severe distribution discrepancies that further shift across illumination levels, we train a unified tokenizer with a shared encoder $E_\phi$, decoder $D_\psi$, and DIJAM in three phases: dataset-adaptive scaling, scale-only pretraining, and DIJAM joint training. The tokenizer uses a convolutional encoder--decoder with residual blocks at each resolution stage; architectural details are given in supplementary Sec.~\suppref{sec:supp-architecture}.

\subsubsection{Phase A: Dataset-Adaptive Scale Estimation}

Motivated by the use of the 99th percentile for linear RAW intensity scaling in RENOIR~\cite{Anaya_2018_RENOIR}, we estimate a single reversible low-light scale for each dataset $\mathcal{D}$ using only its training split. This fixed scale preserves relative intensity differences across illumination levels. Let $q_{0.99}^{(i)}$ denote the 99th percentile of pixel intensities in the $i$-th sampled normalized low-light RAW training patch. Given a target intensity $\tau$, we define
\[
s_{\mathcal{D}} = \frac{\tau}{\mathrm{median}_i\, q_{0.99}^{(i)}}, \quad \tilde{x}_l = s_{\mathcal{D}} x_l,
\]
which scales the low-light branch without hard clipping while leaving the GT branch unchanged.

\subsubsection{Phase B: Scale-Only Tokenizer Pretraining}

We first disable DIJAM and pretrain the shared encoder--decoder to self-reconstruct $\tilde{x}_l$ and $x_g$ using an L1 reconstruction loss in both domains and an additional low-light gradient loss, providing a stable initialization for subsequent affine modulation.

\subsubsection{Phase C: DIJAM Joint Training}

Starting from the scale-only checkpoint, we enable DIJAM across all encoder and decoder stages. The categorical index $d\in\{0,1\}$ takes values $d_l$ and $d_g$ for the scaled low-light and well-exposed RAW domains, respectively. For illumination $L$ expressed numerically in lux, we use
\[
\ell=\log_{10}L+3,
\]
which maps $0.01$, $0.001$, and $0.0001$ lux to $1$, $0$, and $-1$. DIJAM looks up a learned embedding for the domain index $d$ and maps the scalar $\ell$ through an MLP. It then fuses the two embeddings and their elementwise product:
\[
\begin{aligned}
e_d&=\operatorname{Emb}(d), &
e_\ell&=\operatorname{MLP}_\ell(\ell),\\
e_c&=\operatorname{MLP}_c([e_d,e_\ell,e_d\odot e_\ell]).
\end{aligned}
\]
At the $k$-th stage, a zero-initialized projection $P_k$ predicts $(\gamma_k,\beta_k)=P_k(e_c)$ to modulate feature $F_k$ as
\[
F'_k=F_k\odot(1+\gamma_k)+\beta_k,
\]
preserving the initial scale-only mapping at activation. The low-light branch uses its measured $\ell$, whereas the GT branch uses the fixed reference $\ell_{\mathrm{ref}}=0$ so the target latent space remains shared across darkness levels. The tokenizer is trained with
\[
\begin{aligned}
\mathcal{L}_{tok}
={}&\lambda_l
\left(
\|\hat{x}_l-\tilde{x}_l\|_1
+
\lambda_{\nabla}\mathcal{L}_{grad}(\hat{x}_l,\tilde{x}_l)
\right)
\\
&+\lambda_g\|\hat{x}_g-x_g\|_1,
\end{aligned}
\]
by first warming up the domain embedding with the encoder--decoder frozen, then activating the illumination and interaction branches, and finally optimizing the full tokenizer jointly (as verified by the DIJAM ablation in Sec.~\ref{sec:abl-dijam}).

\begin{table*}[!t]
    \centering
    \begingroup
    \setlength{\tabcolsep}{1mm}
    \begin{tabular}{lccc ccc ccc}
        \toprule
        & \multicolumn{3}{c}{$0.01$ lux}
        & \multicolumn{3}{c}{$0.001$ lux}
        & \multicolumn{3}{c}{$0.0001$ lux} \\
        \cmidrule(lr){2-4}\cmidrule(lr){5-7}\cmidrule(lr){8-10}
        Method
        & PSNR $\uparrow$ & SSIM $\uparrow$ & SAM $\downarrow$
        & PSNR $\uparrow$ & SSIM $\uparrow$ & SAM $\downarrow$
        & PSNR $\uparrow$ & SSIM $\uparrow$ & SAM $\downarrow$ \\
        \midrule
        SID       & 22.30& 0.765& 0.065& 21.98& 0.746& 0.068& 20.76& 0.695& 0.075\\
        LLPackNet & 22.15& 0.800& 0.062& 21.81& 0.789& 0.064& 20.99& 0.761& 0.072\\
        RRT       & 22.59& 0.798& 0.073& 22.42& 0.795& 0.074& 21.72& \underline{0.775}& 0.078\\
        DNF       & 21.89& 0.719& 0.094& 21.26& 0.677& 0.098& 20.49& 0.664& 0.106\\
        RAWMamba  & 22.61& 0.773& 0.062& 21.37& 0.737& 0.065& 21.24& 0.723& 0.071\\
        LFM-LLIE  & 23.70 & 0.802 & 0.067 & 21.13 & 0.751 & 0.089 & 19.65 & 0.729 & 0.102 \\
        SIED      & \underline{24.94} & \underline{0.812} & \underline{0.061}
                  & \underline{24.41} & \underline{0.802} & \underline{0.062}
                  & \underline{23.15} & 0.773 & \underline{0.068} \\
        Ours      & \textbf{26.10} & \textbf{0.815} & \textbf{0.058}
                  & \textbf{25.64} & \textbf{0.804} & \textbf{0.060}
                  & \textbf{24.23} & \textbf{0.776} & \textbf{0.064} \\
        \bottomrule
    \end{tabular}
    \endgroup
    \caption{RGB-space quantitative comparison on SIED at three illumination levels. Bold and underlined denote the best and second-best results.}
    \label{tab:sied-quantitative}

    \vspace{\dblfloatsep}
    \includegraphics[width=\textwidth]{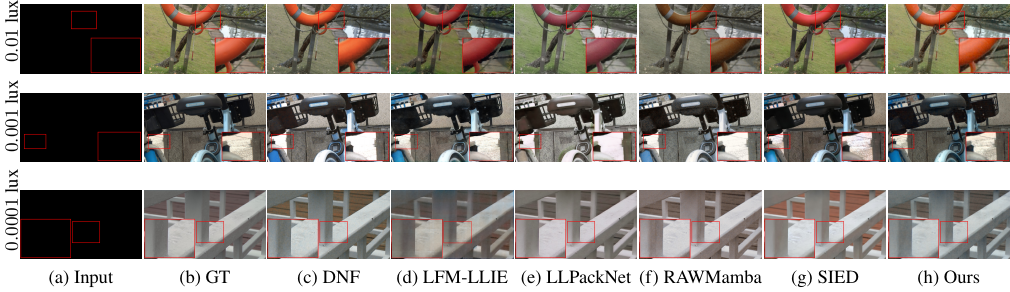}
    \captionof{figure}{Qualitative comparison on SIED without motion blur at $0.01$, $0.001$, and $0.0001$ lux. From left to right: Input, GT, DNF, LFM-LLIE, LLPackNet, RAWMamba, SIED, and Ours. All results are visualized in RGB space under the same evaluation protocol. Red boxes mark corresponding regions, with enlarged views shown in the lower-right corners.}
    \label{fig:sied-qualitative}
\end{table*}

\begin{table*}[!t]
    \centering
    \begingroup
    \setlength{\tabcolsep}{1mm}
    \begin{tabular}{lccc ccc ccc}
        \toprule
        & \multicolumn{3}{c}{$0.01$ lux}
        & \multicolumn{3}{c}{$0.001$ lux}
        & \multicolumn{3}{c}{$0.0001$ lux} \\
        \cmidrule(lr){2-4}\cmidrule(lr){5-7}\cmidrule(lr){8-10}
        Method
        & PSNR $\uparrow$ & SSIM $\uparrow$ & SAM $\downarrow$
        & PSNR $\uparrow$ & SSIM $\uparrow$ & SAM $\downarrow$
        & PSNR $\uparrow$ & SSIM $\uparrow$ & SAM $\downarrow$ \\
        \midrule
        \multicolumn{10}{l}{\textit{Enhancement-only methods}} \\
        \addlinespace[2pt]
        SID       & 18.14& 0.638& 0.084& 17.89& 0.629& 0.087& 17.08& 0.604& 0.092\\
        LLPackNet & 18.34& 0.667& 0.082& 17.87& 0.654& 0.086& 16.61& 0.618& 0.095\\
        RRT       & 17.22& 0.645& 0.098& 17.19& 0.643& 0.104& 17.16& 0.634& 0.115\\
        DNF       & 18.20& 0.622& 0.113& 17.36& 0.565& 0.111& 16.11& 0.571& 0.128\\
        RAWMamba  & 18.56 & 0.611 & 0.114& 17.81 & 0.565 & 0.114& 16.81 & 0.549 & 0.118 \\
        LFM-LLIE  & 19.18 & 0.730 & 0.087 & 18.10 & 0.696 & 0.103 & 16.27 & 0.644 & 0.122 \\
        SIED      & 17.77 & 0.627 & 0.135 & 16.80 & 0.615 & 0.148 & 16.12 & 0.605 & 0.162 \\
        \midrule
        \multicolumn{10}{l}{\textit{Joint enhancement and deblurring methods}} \\
        \addlinespace[2pt]
        DarkIR    & 21.21 & \underline{0.758} & 0.072 & 20.76 & \underline{0.740} & 0.077 & 19.10 & \underline{0.697} & 0.090 \\
        LEDNet    & \underline{21.47} & 0.751 & \underline{0.071} & \underline{21.09} & 0.734 & \underline{0.076} & \underline{19.42} & 0.694 & \underline{0.088} \\
        \midrule
        \multicolumn{10}{l}{\textit{Cascaded enhancement and deblurring methods}} \\
        \addlinespace[2pt]
        SIED $\rightarrow$ EVSSM & 18.17 & 0.654 & 0.128 & 17.19 & 0.630 & 0.140 & 16.42 & 0.608 & 0.154 \\
        RAWMamba $\rightarrow$ EVSSM & 17.33 & 0.630 & 0.083 & 17.29 & 0.617 & 0.086 & 17.26 & 0.590 & 0.093 \\
        \midrule
        Ours      & \textbf{24.15} & \textbf{0.764} & \textbf{0.060}
                  & \textbf{23.61} & \textbf{0.751} & \textbf{0.065}
                  & \textbf{21.66} & \textbf{0.716} & \textbf{0.074} \\
        \bottomrule
    \end{tabular}
    \endgroup
    \caption{RGB-space quantitative comparison on SIDED at three illumination levels. Enhancement-only methods, which do not model camera motion, are trained on blur-preserving enhancement pairs ($x_l^{\mathrm m}\!\to x_g^{\mathrm m}$); joint methods are trained on joint pairs ($x_l^{\mathrm m}\!\to x_g$); the cascades feed the enhancement output to an independently trained EVSSM. All methods are evaluated against the sharp reference $x_g$. Bold and underlined denote the best and second-best results.}
    \label{tab:sided-quantitative}

    \vspace{\dblfloatsep}
    \includegraphics[width=\textwidth]{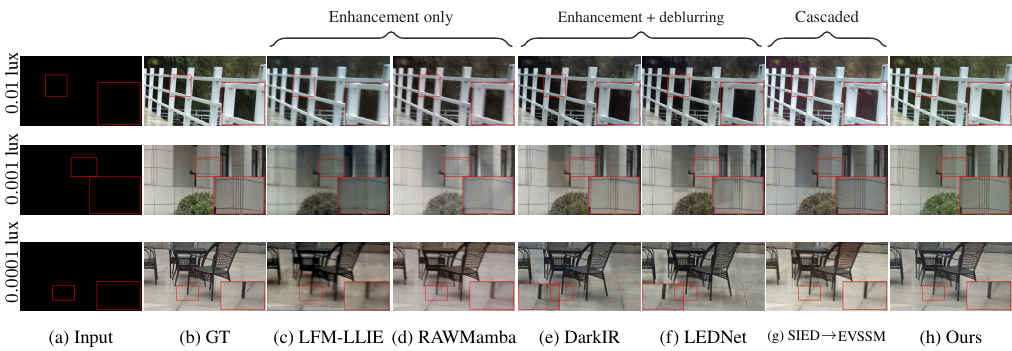}
    \captionof{figure}{Qualitative comparison on SIDED with motion blur at $0.01$, $0.001$, and $0.0001$ lux. We show selected top-performing methods from each baseline category in Table~\ref{tab:sided-quantitative}. From left to right: Input, GT, LFM-LLIE, RAWMamba, DarkIR, LEDNet, SIED $\rightarrow$ EVSSM, and Ours. The displayed restoration results are visualized in RGB space under the same evaluation protocol. Red boxes mark corresponding regions, with enlarged views shown in the lower-right corners. Zoom in to see more details.}
    \label{fig:sided-qualitative}
\end{table*}

\subsection{Stage II: Unified Conditional MeanFlow}

With the Stage I tokenizer frozen, each aligned scene yields four latents corresponding to low-light sharp, low-light blurred, normal-exposure sharp, and normal-exposure blurred RAW observations:
\[
\begin{array}{ll}
z_{ls}=E_\phi(s_{\mathcal D}x_l;d_l,\ell), & z_{lb}=E_\phi(s_{\mathcal D}x_l^{\mathrm m};d_l,\ell),\\
z_{gs}=E_\phi(x_g;d_g,\ell_{\mathrm{ref}}), & z_{gb}=E_\phi(x_g^{\mathrm m};d_g,\ell_{\mathrm{ref}}).
\end{array}
\]

\subsubsection{Shared Backbone and Conditional Residual Heads}

For source and target latents $(z_{\mathrm{in}},z_{\mathrm{tar}})$, MeanFlow defines
\[
z_t=(1-t)z_{\mathrm{tar}}+tz_{\mathrm{in}},\qquad
v=z_{\mathrm{in}}-z_{\mathrm{tar}}.
\]
The shared backbone $F_\theta$ takes the interpolated latent $z_t$, source latent $z_{\mathrm{in}}$, time variables $(t,t-r)$, and MeanFlow condition $c_{\mathrm{mf}}$. In a single forward pass, the backbone's decoder produces a shared feature map, which two lightweight convolutional heads use to predict the enhancement and deblurring residuals (as verified by the dual-head ablation in Sec.~\ref{sec:abl-heads}):
\[
(u_{\mathrm{enh}},u_{\mathrm{deb}})
=F_\theta(z_t,z_{\mathrm{in}},t,t-r,c_{\mathrm{mf}}).
\]
We combine the two residuals into the average velocity:
\[
u_\theta=u_{\mathrm{enh}}+b\alpha u_{\mathrm{deb}}.
\]
Using the Jacobian--vector product for the directional derivative $\dot u_\theta$, the MeanFlow target and objective are
\[
u_{\mathrm{tgt}}=v-(t-r)\dot u_\theta, \quad
\mathcal{L}_{mf}
=\mathrm{E}_{t,r}
\left[
w\left\|u_\theta-\operatorname{sg}(u_{\mathrm{tgt}})\right\|_2^2
\right],
\]
where $w$ is a detached adaptive normalization weight and $\operatorname{sg}$ denotes stop-gradient.

\subsubsection{Task-Conditioned Velocity Decomposition}

For joint restoration, we encode the progressive RAW target
\[
x_g^{(\alpha)}=(1-\alpha)x_g^{\mathrm m}+\alpha x_g,
\qquad z_\alpha=E_\phi(x_g^{(\alpha)};d_g,\ell_{\mathrm{ref}}).
\]
At the endpoint $(r,t)=(0,1)$, the joint latent displacement decomposes through the blurred normal-exposure latent $z_{gb}$ as
\[
z_{lb}-z_\alpha=(z_{lb}-z_{gb})+(z_{gb}-z_\alpha),
\]
which motivates our dual-head design: $u_{\mathrm{enh}}$ models low-light exposure and appearance recovery, while $u_{\mathrm{deb}}$ supplies the motion-deblurring correction when gated by $b\alpha$ (as verified by the deblurring-output ablation in Sec.~\ref{sec:abl-deblur}), with $\mathcal{L}_{mf}$ supervising their combined velocity $u_\theta$.

\subsubsection{Four-Phase Curriculum}

Phase A sets $\alpha=0$ and trains on equal proportions of clean enhancement pairs $(z_{ls},z_{gs})$ and blur-preserving enhancement pairs $(z_{lb},z_{gb})$. Phase B adds deblurring on well-exposed pairs $(z_{gb},z_{gs})$, activating the deblurring contribution where structural cues have higher SNR. In Phase C, with training progress $q\in[0,1]$ and schedule
\[
A(q)=\sin^2\!\left(\frac{\pi q}{2}\right),
\]
half of the joint samples use $\alpha=A(q)$ and the rest sample $\alpha\sim\mathcal{U}[0,A(q)]$. Phases A--C are trained solely with $\mathcal L_{mf}$. Phase D fixes $\alpha=1$ for joint restoration while retaining clean enhancement samples to prevent exposure drift and over-sharpening, switching supervision to Retinex and RGB-space losses (as verified by the Retinex ablation in Sec.~\ref{sec:abl-retinex}).

\subsubsection{Training Objective}

Given the one-step endpoint prediction $\hat z=z_{\mathrm{in}}-u_\theta(z_{\mathrm{in}},z_{\mathrm{in}},1,1,c_{\mathrm{mf}})$, the frozen GT decoder and frozen differentiable ISP produce $\hat x_g=D_\psi(\hat z;d_g,\ell_{\mathrm{ref}})$ and $\hat y=\operatorname{ISP}(\hat x_g)$, with RGB target $y_g=\operatorname{ISP}(x_g)$. Using a DecomNet \cite{Yi_2025_TPAMI} pretrained on well-exposed ground-truth RGB images from the training set and then frozen, we extract reflectance and illumination pairs $(\hat R,\hat I)$ and $(R_g,I_g)$ from $\hat y$ and $y_g$ to define the Retinex and RGB losses:
\[
\mathcal L_{ret}
=\|\hat R-R_g\|_1+\lambda_I\|\hat I-I_g\|_1,
\]
\[
\begin{aligned}
\mathcal L_{mse}^{rgb}&=\|\hat y-y_g\|_2^2,\\
\mathcal L_{sam}^{rgb}
&=\arccos\!\left(
\frac{\langle\hat y,y_g\rangle}
{\|\hat y\|_2\|y_g\|_2+\epsilon}
\right).
\end{aligned}
\]
The complete phase-dependent training objective is
\[
\mathcal L=
\begin{cases}
\mathcal L_{mf}, & \text{Phases A--C},\\[2pt]
\begin{aligned}
&\lambda_{ret}\mathcal L_{ret}
+\lambda_{mse}\mathcal L_{mse}^{rgb}
\\
&\quad+\lambda_{sam}\mathcal L_{sam}^{rgb}
\end{aligned}, & \text{Phase D}.
\end{cases}
\]
Throughout Phase D, the tokenizer, ISP, and DecomNet remain frozen, and gradients update only the shared MeanFlow backbone and its two residual heads.

\subsection{One-Step Conditional Inference}

For low-light inference, the full condition $c$ supplies $c_{\mathrm{tok}}=(d_l,\ell)$ for encoding the scaled input:
\[
z_{\mathrm{in}}=E_\phi(s_{\mathcal D}x_{\mathrm{in}};d_l,\ell).
\]
We then use $c_{\mathrm{mf}}$ for one endpoint evaluation at $(t,t-r)=(1,1)$:
\[
\begin{array}{c}
(u_{\mathrm{enh}},u_{\mathrm{deb}})
=F_\theta(z_{\mathrm{in}},z_{\mathrm{in}},1,1,c_{\mathrm{mf}}),\\[2pt]
u=u_{\mathrm{enh}}+b\alpha u_{\mathrm{deb}},\qquad
\hat z_g=z_{\mathrm{in}}-u,\\[2pt]
\hat x_g=D_\psi(\hat z_g;d_g,\ell_{\mathrm{ref}}),\qquad
\hat y_g=\mathrm{ISP}(\hat x_g).
\end{array}
\]
Clean SIED inputs use $b=\alpha=0$ and motion-degraded SIDED inputs use $b=\alpha=1$, requiring only a single MeanFlow backbone evaluation with shared parameters.

%% file: sec/5_experiments.tex
\section{Experiments}

\subsection{Experimental Setup}

\paragraph{Datasets and protocol.}
We evaluate on SIED and SIDED with the Sony sensor at $0.01$, $0.001$, and $0.0001$ lux. Both benchmarks share 1,500 training and 180 test scenes at each illumination level. RAW and RGB models take RAW and ISP-rendered inputs, respectively, and one frozen MetaISP~\cite{Chen_2024_MetaISP} performs all RAW-to-RGB rendering, including within cascades. We report RGB PSNR, SSIM, and SAM~\cite{Yang_2026_GZAP} against shared ISP-rendered sharp references. A single LEAD-MF checkpoint, trained once on all SIDED settings, serves both benchmarks and all levels ($b=\alpha=0$ for SIED and $b=\alpha=1$ for SIDED). All experiments use an NVIDIA GeForce RTX 4090.

\paragraph{Compared methods.}
We compare against three baseline categories: (1) enhancement-only methods, namely the direct-regression SID \cite{Chen_2018_CVPR}, LLPackNet \cite{lamba2020fastlightweightrestorationdark}, RRT \cite{Lamba_2021_CVPR}, DNF \cite{Jin_2023_CVPR}, and RAWMamba \cite{Chen_2026}, the flow-matching LFM-LLIE \cite{Ciubotariu_2026_CVPRW}, and the diffusion-based SIED \cite{Jiang_2025_ICCV}; (2) RGB-domain joint enhancement and deblurring methods, DarkIR \cite{Feijoo_2025_DarkIR} and LEDNet \cite{Zhou_2022_LEDNet}; and (3) the cascades SIED $\rightarrow$ EVSSM and RAWMamba $\rightarrow$ EVSSM, where EVSSM is an RGB image deblurring model \cite{Kong_2025_EVSSM}, trained on SIDED's blurred/sharp normal-exposure pairs.

\subsection{Comparison with Existing Methods}

\paragraph{Results on SIED.}
LEAD-MF ranks first in all metrics at every level (Table~\ref{tab:sied-quantitative}), outperforming the strongest baseline SIED by $1.16$, $1.23$, and $1.08$ dB PSNR from $0.01$ to $0.0001$ lux. In Fig.~\ref{fig:sied-qualitative}, LEAD-MF recovers sharper structures, \eg, the bicycle boundary, mechanical details, and railings, with more faithful colors and less noise.

\paragraph{Results on SIDED.}
In Table~\ref{tab:sided-quantitative}, enhancement-only methods retain motion blur. At $0.01$ lux, adding EVSSM gains only $0.40$ dB over SIED and loses $1.23$ dB with RAWMamba: the residual noise and artifacts of extreme-darkness enhancement exceed what a separate RGB deblurring stage can reliably correct. RGB-domain joint methods also remain limited by ISP information loss. These results motivate unified RAW enhancement and deblurring: LEAD-MF is best in all metrics at every level, surpassing the strongest baseline LEDNet by $2.68$, $2.52$, and $2.24$ dB PSNR. In Fig.~\ref{fig:sided-qualitative}, competing methods leave smearing on fence bars, shrubs, chair frames, and floor seams, whereas LEAD-MF restores sharp boundaries and accurate colors. The remaining baselines are shown in supplementary Fig.~\suppref{fig:supp-additional-qualitative} (Sec.~\suppref{sec:supp-additional-qualitative}).

\paragraph{Efficiency.}
On $0.01$-lux SIDED, one-step LEAD-MF achieves the highest RGB PSNR ($24.15$ dB) in $1164$ ms, gaining $6.38$ and $4.97$ dB over SIED and LFM-LLIE while reducing inference time by $16.9\%$ and $83.0\%$, respectively; it is also more accurate and faster than both cascades (supplementary Fig.~\suppref{fig:supp-latency-psnr} and Sec.~\suppref{sec:supp-efficiency}).

\subsection{Ablation Studies}
\label{sec:ablation}

\begin{table}[t]
    \centering
    \small
    \setlength{\tabcolsep}{3pt}
    \begin{tabular}{@{}llccc@{}}
        \toprule
        & Variant & PSNR $\uparrow$ & SSIM $\uparrow$ & SAM $\downarrow$ \\
        \midrule
        (a) & w/o DIJAM & 46.24 & 0.910 & 0.075 \\
            & w/ DIJAM & \textbf{50.23} & \textbf{0.933} & \textbf{0.058} \\
        \midrule
        (b) & w/o $\mathcal{L}_{ret}$ & 22.73 & 0.757 & 0.066 \\
            & w/ $\mathcal{L}_{ret}$ & \textbf{24.15} & \textbf{0.764} & \textbf{0.060} \\
        \midrule
        (c) & Single head, clean & 23.60 & 0.801 & 0.060 \\
            & Two heads, clean & \textbf{26.10} & \textbf{0.815} & \textbf{0.058} \\
            & Single head, joint & 22.09 & 0.713 & 0.071 \\
            & Two heads, joint & \textbf{24.15} & \textbf{0.764} & \textbf{0.060} \\
        \midrule
        (d) & Head off vs.\ $x_g$ & 22.26 & 0.724 & 0.061 \\
            & Full vs.\ $x_g$ & \textbf{24.15} & \textbf{0.764} & \textbf{0.060} \\
            & Head off vs.\ $x_g^{\mathrm m}$ & 27.84 & 0.928 & 0.047 \\
        \bottomrule
    \end{tabular}
    \caption{Ablations in RGB space: (a) tokenizer self-reconstruction; (b) Retinex loss on SIDED joint restoration; (c) single- vs.\ two-head models on SIED clean ($b=0$) and SIDED joint ($b=\alpha=1$) restoration; (d) inference-time removal of $u_{\mathrm{deb}}$ on 180 SIDED scenes ($b=\alpha=1$). (b)--(d) use $0.01$ lux. Protocols are detailed in supplementary Sec.~\suppref{sec:supp-ablations}.}
    \label{tab:ablation}
\end{table}

\subsubsection{Effect of DIJAM}
\label{sec:abl-dijam}
In standalone tokenizer self-reconstruction with identical scaling, DIJAM gains $3.99$ dB PSNR and reduces SAM by $22.7\%$ (Table~\ref{tab:ablation}(a)), confirming that joint domain--illumination modulation bridges the gap across exposure domains and darkness levels.

\subsubsection{Effect of Retinex Consistency}
\label{sec:abl-retinex}
Toggling only $\lambda_{ret}$ in Stage~II Phase~D, Retinex consistency adds $1.42$ dB PSNR and reduces SAM by $9.1\%$ on $0.01$-lux SIDED joint restoration (Table~\ref{tab:ablation}(b)), complementing the RGB and spectral losses.

\subsubsection{Dual-Head Decomposition}
\label{sec:abl-heads}
A single-head variant predicting the combined velocity under the same curriculum is $2.50$ and $2.06$ dB lower in PSNR on $0.01$-lux clean and joint restoration (Table~\ref{tab:ablation}(c)). Its joint result is even below the head-off output of LEAD-MF ($22.09$ vs.\ $22.26$ dB, Sec.~\ref{sec:abl-deblur}), suggesting interference between enhancement and deblurring that the decoupled residuals avoid.

\subsubsection{Deblurring-Output Intervention}
\label{sec:abl-deblur}
Removing $u_{\mathrm{deb}}$ from the same checkpoint at inference yields an output close to the blurred reference $x_g^{\mathrm m}$, whereas the full prediction gains $1.89$ dB PSNR against the sharp $x_g$ (Table~\ref{tab:ablation}(d)), confirming that $u_{\mathrm{deb}}$ captures the motion-deblurring residual.

%% file: sec/6_conclusion.tex
\section{Conclusion}

We presented LEAD-MF, a unified RAW restoration framework for extremely dark handheld imaging where severe illumination loss is frequently accompanied by motion blur, together with the paired SIDED benchmark for controlled evaluation. By equipping a shared RAW tokenizer with DIJAM, LEAD-MF aligns low-light and well-exposed RAW observations across illumination levels into a compact latent space for full-image processing. Within this space, a task-conditioned MeanFlow backbone predicts and composes enhancement and deblurring velocity residuals in a single step, further guided by training-time Retinex consistency. Extensive experiments from $0.01$ to $0.0001$ lux on SIED and SIDED show that LEAD-MF consistently surpasses enhancement-only, joint, and cascaded baselines across pixel fidelity (PSNR), structural quality (SSIM), and color consistency (SAM). Future work will extend the framework to diverse camera sensors and more dynamic scenes.

%% file: supplementary.tex
\renewcommand{\dblfloatpagefraction}{0.90}
\appendix
\setcounter{figure}{0}
\setcounter{table}{0}
\setcounter{equation}{0}
\renewcommand{\theHfigure}{supp.\arabic{figure}}
\renewcommand{\theHtable}{supp.\arabic{table}}
\renewcommand{\theHequation}{supp.\arabic{equation}}
\maketitlesupplementary

\input{supp/A_dataset_construction}
\input{supp/B_architecture_training}

\input{supp/C_baseline_reproduction}
\input{supp/D_additional_ablations}
\input{supp/E_efficiency}
\input{supp/F_additional_qualitative}
\FloatBarrier

{\small
\bibliographystyle{ieeenat_fullname}
\bibliography{main}
}

%% file: supp/A_dataset_construction.tex
\section{Additional Details of SIDED Construction}
\label{sec:supp-sided}

\paragraph{Source signal and construction order.}
SIDED builds on the paired scenes captured with the Sony sensor in SIED~\cite{Jiang_2025_ICCV} and adds motion-degraded observations for each scene. SIED first aligns a captured low-light RAW image $I_{\mathrm{cap}}$ to a standard image $I_{\mathrm{st}}$ through
\begin{equation}
I_{\mathrm{syn}}=I_{\mathrm{cap}}
\left(\frac{\operatorname{Expo}(I_{\mathrm{st}})}
{\operatorname{Expo}(I_{\mathrm{cap}})}+\eta\right),
\label{eq:supp-isyn}
\end{equation}
where $\operatorname{Expo}$ averages the Bayer channels and $\eta$ is SIED's illumination-adjustment factor. SIED then adds ISO-calibrated noise. We obtained the original noise-generation seeds from the SIED authors and replayed their synthesis with the corresponding scene and camera settings. Denoting the replayed noise \emph{realization} by $N_{\mathrm{all}}^{(0)}$, we remove SIED's added noise as
\begin{equation}
I_{\mathrm{syn}}=x_l^{\mathrm{SIED}}-N_{\mathrm{all}}^{(0)},
\qquad
N_{\mathrm{all}}^{(0)}=N_P^{(0)}+N_G^{(0)}+N_D^{(0)}.
\label{eq:supp-noise-removal}
\end{equation}
Here $N_P$, $N_G$, and $N_D$ denote the signal-dependent Poisson, Gaussian, and sampled dark-frame components of SIED's calibrated model. Equation~\eqref{eq:supp-noise-removal} subtracts the exactly replayed realization, recovering the noise-free signal on which our motion synthesis operates. We also measure a dark-frame noise map $D_{\mathrm{ref}}$ using a sensor of the same model and subtract it from $I_{\mathrm{syn}}$ before motion synthesis:
\begin{equation}
I_{\mathrm{syn}}^{\mathrm{df}}=I_{\mathrm{syn}}-D_{\mathrm{ref}}.
\label{eq:supp-dark-frame-correction}
\end{equation}
This dark-frame correction precedes the motion operator and is distinct from removing the replayed SIED noise realization above. The subsequent construction is
\begin{equation}
I_{\mathrm{syn}}^{\mathrm{df}}
\xrightarrow{\text{camera-motion integration}}
\mathcal B_p(I_{\mathrm{syn}}^{\mathrm{df}})
\xrightarrow{\text{SIED calibrated noise}}
x_l^{\mathrm m}.
\end{equation}
This order keeps the newly sampled sensor noise outside the motion operator. The original SIED pair is retained as the blur-free counterpart.

\paragraph{Random camera-motion trajectories.}
The SIDED motion generator uses the stochastic spatial-trajectory update of Boracchi and Foi~\cite{Boracchi_2012_TIP}, followed by our variable-speed exposure-time reparameterization. Let $p_t\in\mathbb R^2$ and $v_t\in\mathbb R^2$ be the position and displacement vector at geometric sample $t$, with $M$ samples, path length $L$, step length $\delta=L/(M-1)$, and perturbation strength $a$. These geometric samples are subsequently mapped to exposure time. For each path, we draw $c=0.7U_c$, $g=10U_g$, and $f=0.2U_f$, where all $U$ variables are uniform on $[0,1]$. The spatial update is
\begin{equation}
\begin{aligned}
J_t &= 2\,\mathbf 1[U_t<fa]\,R(\pi+U'_t-0.5)v_t,\\
\tilde v_{t+1} &= v_t+J_t+a\delta(g\epsilon_t-cp_t),\\
v_{t+1} &= \delta\,\frac{\tilde v_{t+1}}{\|\tilde v_{t+1}\|_2},
\qquad p_{t+1}=p_t+v_{t+1},
\end{aligned}
\label{eq:supp-trajectory}
\end{equation}
where $\epsilon_t\sim\mathcal N(0,I_2)$, $R(\cdot)$ is a planar rotation, and $U_t,U'_t$ are fresh uniform draws. The initial direction is uniform over $[0,2\pi)$. The impulse $J_t$ changes direction occasionally, while normalization fixes the geometric step length. This update specifies a spatial curve with constant geometric sampling density; the traversal speed during exposure is defined separately below. The stochastic update thus yields diverse, realistic shake trajectories with smooth drift and occasional abrupt direction changes.

\paragraph{Variable-speed exposure-time sampling.}
To introduce acceleration and deceleration along the spatial curve, we sample $K=8$ log-speed controls $a_k\sim\mathcal U(-\log 2,\log 2)$ at uniform knots $u_k=k/(K-1)$ on normalized exposure time $u\in[0,1]$. Between adjacent knots, we interpolate log-speed using the raised-cosine weight $w(\xi)=(1-\cos(\pi\xi))/2$, where $\xi=(u-u_k)/(u_{k+1}-u_k)$. This gives $a(u)=(1-w(\xi))a_k+w(\xi)a_{k+1}$ and a smooth positive speed weight $q(u)=\exp(a(u))$. For interval midpoints $\bar u_j=(j+1/2)/(M-1)$, normalized arclength progress is
\begin{equation}
s_m=\frac{\sum_{j=0}^{m-1}q(\bar u_j)}{\sum_{j=0}^{M-2}q(\bar u_j)},
\qquad m=0,\ldots,M-1.
\label{eq:supp-time-reparameterization}
\end{equation}
We evaluate the piecewise-linear spatial curve at arclength $Ls_m$ to obtain equal-exposure-time positions $\bar p_m$, and center them by their temporal mean, $\tilde p_m=\bar p_m-M^{-1}\sum_n\bar p_n$. Normalization preserves traversal of the complete geometric curve, while relative speed changes continuously over the exposure. Equal-time integration therefore assigns greater dwell weight to slower portions of the path. The unnormalized speed weights lie in $[1/2,2]$; actual relative speeds are obtained by dividing by their exposure-time mean. This temporal law extends the Boracchi--Foi model with non-uniform traversal speed. Because speeds are defined on normalized exposure time, the same trajectory model applies to any exposure duration.

\paragraph{PSF rasterization and Bayer RAW blur.}
We convert the centered, equal-time path samples $\tilde p_t$ into a unit-mass point-spread function (PSF) by bilinear splatting. On a packed Bayer plane, the PSF is
\begin{equation}
\begin{aligned}
h_p(i,j)&=\frac{1}{M}\sum_{t=0}^{M-1}
\phi(i-\tilde p_{t,x}/2)\\
&\qquad\qquad\cdot\phi(j-\tilde p_{t,y}/2),\\
\phi(s)&=\max(1-|s|,0),
\end{aligned}
\label{eq:supp-psf}
\end{equation}
for pixel coordinates $(i,j)$ centered on the kernel grid. The factor $1/2$ converts original mosaic-pixel displacement to packed-plane displacement. Each of the four same-color CFA planes is convolved separately with $h_p$, avoiding interpolation between different CFA colors. Convolution uses reflection padding; the result is rounded and stored as unsigned 16-bit RAW. The generator uses 2,000 equal-time samples, a 60-pixel geometric path length, and a perturbation strength of 0.005. The default minimum support is $61\times61$ in original RAW coordinates and $31\times31$ on each packed CFA plane. If exposure-centroid centering requires more space, the odd-sized support is enlarged to retain the complete trajectory without clipping its PSF mass.

\paragraph{Calibrated noise and paired records.}
Let $S_m=\mathcal B_p(I_{\mathrm{syn}}^{\mathrm{df}})$ be the motion-integrated RAW signal. After motion integration, we draw a \emph{new} calibrated-noise realization with SIED's camera- and ISO-dependent parameters $\kappa$:
\begin{equation}
\begin{aligned}
x_l^{\mathrm m}={}&S_m+N_P(S_m;\kappa,\xi_P)
 +N_G(\kappa,\xi_G)\\
&+N_D(\kappa,\xi_D).
\end{aligned}
\label{eq:supp-post-motion-noise}
\end{equation}
In centered form, the SIED Poisson--Gaussian components can be written as
\begin{equation}
\begin{aligned}
N_P(S;\kappa)&=k_\kappa\operatorname{Pois}(S/k_\kappa)-S,\\
N_G(\kappa)&\sim\mathcal N(0,\sigma_\kappa^2),
\qquad N_D(\kappa)\sim\mathcal D_\kappa,
\end{aligned}
\label{eq:supp-sied-noise-components}
\end{equation}
where $k_\kappa$ and $\sigma_\kappa^2$ are SIED's ISO-dependent calibrated parameters and $\mathcal D_\kappa$ is its dark-frame database at the corresponding ISO. The Poisson component depends on the motion-integrated signal, whereas the Gaussian and dark-frame components use the original camera calibration. The new random draws $\xi_P,\xi_G,\xi_D$ are independent of the replayed realization in Eq.~\eqref{eq:supp-noise-removal}. We preserve the camera metadata of the corresponding source scene. The well-exposed motion-degraded observation is obtained with the same trajectory, $x_g^{\mathrm m}=\mathcal B_p(x_g)$, so that the two motion-degraded observations remain spatially paired, while the original sharp well-exposed image $x_g$ remains the restoration reference.

\paragraph{Dataset statistics.}
SIDED shares the Sony-sensor scene split of SIED: at each of the three
illumination levels ($0.01$, $0.001$, and $0.0001$ lux), 1,500 scenes are
used for training and 180 for testing, with identical scene IDs in SIED and
SIDED. Each scene provides the four aligned RAW observations
$(x_l,x_g,x_l^{\mathrm m},x_g^{\mathrm m})$, so the clean and
motion-degraded evaluations are performed on exactly the same scenes and
sharp references.

\begin{figure*}[t]
    \centering
    \includegraphics[width=0.96\textwidth]{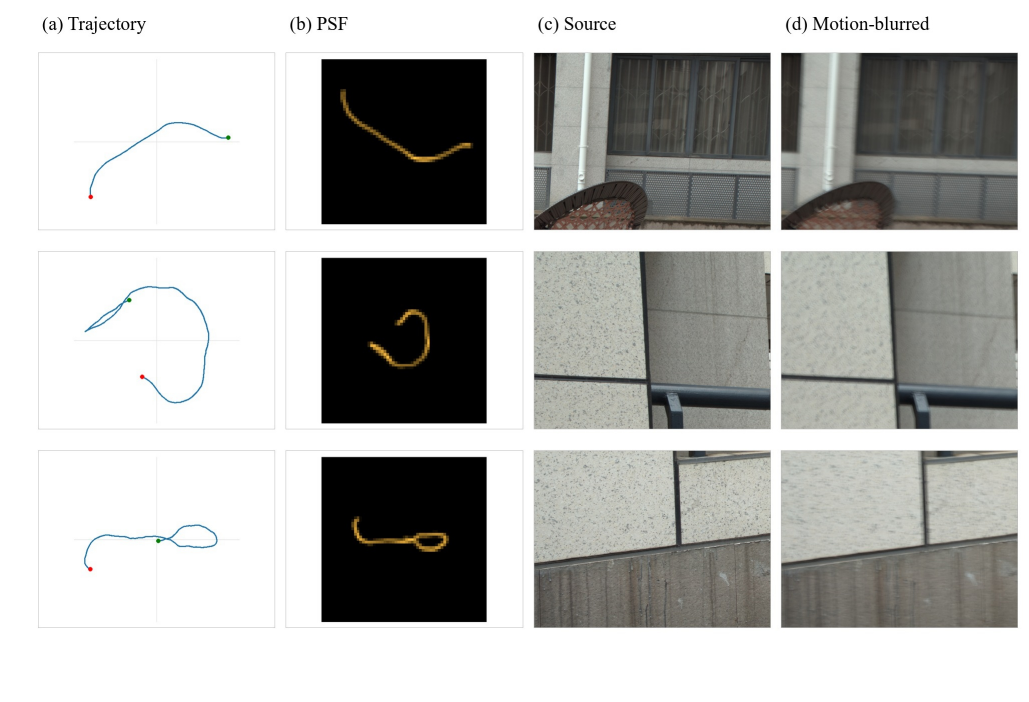}
    \caption{Illustrative variable-speed motion synthesis. Each row shows
    a different sampled camera trajectory (green start, red end), its
    $61\times61$ RAW-domain point-spread function, and the source and
    motion-blurred RAW images rendered with the same ISP. Source and blurred
    previews use identical crops. These panels visualize motion integration
    before the separate calibrated-noise sampling used for the final SIDED
    observations.}
    \label{fig:supp-motion-previews}
\end{figure*}

\paragraph{Motion-synthesis previews.}
Figure~\ref{fig:supp-motion-previews} shows how three sampled camera paths produce distinct blur patterns. The enlarged crops make their effect on chair lattices, wall texture, and straight edges visible. These previews isolate the motion stage; in the final SIDED observations, calibrated noise is sampled afterwards.

%% file: supp/B_architecture_training.tex
\section{Architecture and Training Details}
\label{sec:supp-architecture}

\paragraph{RAW tokenizer.}
The tokenizer maps a four-channel packed Bayer RAW image to a continuous
latent and reconstructs packed RAW from that latent. Its encoder and decoder
share one set of convolutional weights across the scaled low-light and
well-exposed domains. Each resolution stage consists of two-convolution
residual blocks; the encoder reduces spatial resolution and the decoder
reverses this transformation. A $1\!\times\!1$ projection maps the encoder
feature to 32 latent channels, and a second $1\!\times\!1$ projection maps
the latent back to the decoder feature width. The configuration uses a
fourfold spatial compression.
The same tokenizer architecture is used at all three illumination levels.

The tokenizer receives a domain indicator $d$ and a continuous illumination
value $\ell$, as defined in the main paper. DIJAM embeds these two inputs
separately and includes their elementwise interaction in the conditioning
vector. At the end of each encoder and decoder resolution stage, after its
residual blocks and before the next resolution change, a stage-specific
projection converts this vector into channel-wise scale and bias parameters.
The modulation is applied once per stage, rather than after every
convolution. Its output has the same spatial and channel dimensions as the
unmodulated feature. The well-exposed branch uses the fixed reference
illumination $\ell_{\mathrm{ref}}=0$; the low-light branch uses the
illumination of the observation.
Thus, illumination changes the feature modulation without introducing a
separate encoder or decoder for each light level.

\paragraph{Conditional MeanFlow network.}
The tokenizer remains frozen while training the restoration network. The
MeanFlow backbone is a single encoder--decoder U-Net operating on tokenizer
latents. The interpolated latent $z_t$ and the source latent $z_{\rm in}$
pass through separate shallow convolutions; their resulting features are
concatenated before the shared U-Net. The time variables $t$ and $t-r$ use
positional embeddings. The MeanFlow condition $c_{\mathrm{mf}}=(e,\ell,b,\alpha)$
contains the source-exposure indicator ($e=0$ for low-light and $e=1$ for
well-exposed sources), normalized illumination, the blur indicator, and the
deblurring strength. We project these controls into the time embedding, which
supplies channel-wise scale and shift parameters to the residual blocks. Skip connections convey
encoder features to the matching decoder resolutions. Two output
convolutions read the \emph{same} final decoder feature to predict
$u_{\rm enh}$ and $u_{\rm deb}$; they do not require two U-Net evaluations.
The combined prediction is $u_{\rm enh}+b\alpha u_{\rm deb}$.

\begin{table}[t]
\centering
\small
\setlength{\tabcolsep}{3pt}
\begin{tabular}{p{0.21\columnwidth}p{0.31\columnwidth}p{0.36\columnwidth}}
\toprule
Module & Input & Output / role \\
\midrule
Tokenizer encoder & Packed RAW, $(d,\ell)$ & 32-channel continuous latent \\
Tokenizer decoder & Latent, selected $(d,\ell)$ & Four-channel packed RAW \\
MeanFlow stems & $z_t$, $z_{\rm in}$ & Fused U-Net input feature \\
Condition projection & $(t,t-r,e,\ell,b,\alpha)$ & Residual-block affine modulation \\
Parallel heads & Shared decoder feature & $u_{\rm enh}$, $u_{\rm deb}$ \\
\bottomrule
\end{tabular}
\caption{Information flow of the shared tokenizer and conditional MeanFlow
backbone. The tokenizer and MeanFlow receive different subsets of the
structured condition.}
\label{tab:supp-architecture}
\end{table}

\paragraph{Tokenizer training.}
The dataset-specific low-light scale is estimated from the training split
with target intensity $\tau=0.5$ and fixed thereafter. The tokenizer is
first trained for self-reconstruction
of scaled low-light and well-exposed RAW without DIJAM. Starting from this
scale-only model, the domain modulation is initialized near the identity.
Its domain component is warmed up before the illumination and interaction
components are activated, after which the complete tokenizer is trained
jointly. The two RAW domains share the encoder and decoder throughout. The
reconstruction and low-light gradient objectives are specified in the main
paper. Both scale-only pretraining and DIJAM joint training use
$\lambda_l=2.0$, $\lambda_g=1.0$, and $\lambda_{\nabla}=0.2$ in
$\mathcal L_{tok}$; the well-exposed branch has no gradient-loss term
(weight zero). No restoration target is used to train the tokenizer at
this stage.

\paragraph{One model for all settings.}
Every LEAD-MF result reported in the main paper and in this supplementary
material is produced by a single set of weights, \ie, one tokenizer and one
MeanFlow network. This model is trained once on the full SIDED training
split, covering all three illumination levels ($0.01$, $0.001$, and
$0.0001$ lux) and all four observations of each scene: the clean pair
$(x_l,x_g)$, which coincides with SIED, and the motion-degraded pair
$(x_l^{\rm m},x_g^{\rm m})$. The same checkpoint performs clean enhancement
on SIED and joint enhancement--deblurring on SIDED at every illumination
level; only the condition $(\ell,b,\alpha)$ changes at inference, and no
benchmark-, level-, or task-specific fine-tuning is applied. The ablation
variants are trained under the identical setting. In contrast, every
baseline is trained separately for each benchmark, and the joint methods
additionally for each illumination level (Sec.~\ref{sec:supp-baselines}).

\paragraph{Restoration curriculum and supervision.}
After freezing the tokenizer, we train the conditional MeanFlow network in
four phases. Phase A introduces enhancement with the deblurring contribution
gated off. Phase B adds the well-exposed deblurring task, allowing the motion
component to learn from higher-SNR inputs. Phase C introduces joint
restoration and progressively expands the sampled range of $\alpha$.
Phases A--C use the combined-velocity MeanFlow objective $\mathcal L_{mf}$;
there is no independent endpoint loss assigned to either output head.
Phase D fixes $\alpha=1$ for joint samples, retains clean enhancement
samples, and optimizes the Retinex, RGB MSE, and RGB SAM terms specified in
the main paper, with $\mathcal L_{mf}$ disabled. We use
$\lambda_{ret}=0.1$, $\lambda_{mse}=1.0$, and $\lambda_{sam}=0.1$;
within $\mathcal L_{ret}$, the illumination weight is $\lambda_I=1.0$.
The Retinex decomposition
network is pretrained separately on well-exposed ground-truth RGB images
from the training set and, during Phase D, receives the predicted and
reference RGB images rendered by MetaISP~\cite{Chen_2024_MetaISP}. The tokenizer, differentiable ISP,
and decomposition network are frozen; gradients propagate through them to
update only the MeanFlow backbone and its heads.
The two heads operate in parallel in every phase; the curriculum changes
their task exposure, not the order in which an image is restored.

\paragraph{Time sampling.}
Following the logit-normal sampling strategy of MeanFlow~\cite{NEURIPS2025_6d13e085},
for each sample in Phases A--C we draw independent
$\epsilon_1,\epsilon_2\sim\mathcal N(0,1)$ and set
$a_j=\sigma(\epsilon_j-0.4)$ for $j\in\{1,2\}$, where $\sigma$ is the sigmoid.
We then sample
\begin{equation}
(r,t)=
\begin{cases}
(a_1,a_1), & \text{prob. }0.75,\\
(\min(a_1,a_2),\max(a_1,a_2)), & \text{prob. }0.25.
\end{cases}
\label{eq:supp-time-sampling}
\end{equation}
The branch is chosen independently for each sample in the batch and
independently of the Gaussian draws. When $r=t$, the correction
$(t-r)\dot u_\theta$ vanishes and $u_{\mathrm{tgt}}=v$.
Phase D instead uses the fixed endpoint $(r,t)=(0,1)$.

\paragraph{Optimizer and MeanFlow loss.}
Stage II uses AdamW with $(\beta_1,\beta_2)=(0.9,0.95)$ and weight decay
$0.01$. In Phases A--C, the JVP-based target $u_{\mathrm{tgt}}$ from the
main paper is used in
\begin{equation}
\begin{aligned}
e_i &= \frac{1}{CHW}
\left\|u_{\theta,i}-\operatorname{sg}(u_{\mathrm{tgt},i})\right\|_2^2,\\
w_i &= \left(\operatorname{sg}(e_i)+10^{-3}\right)^{-1/2},\\
\mathcal L_{mf} &= \frac{1}{B}\sum_{i=1}^{B}w_i e_i.
\end{aligned}
\label{eq:supp-mf-normalization}
\end{equation}
Here $CHW$ is the latent element count and $B$ the batch size: errors are
averaged per sample before weighting. The target and $w_i$ are detached;
gradients flow through $e_i$. In the main paper's squared-norm notation,
$w=w_i/(CHW)$ for sample $i$.

For the model used in the reported experiments, the four MeanFlow training
phases use the configurations in Table~\ref{tab:supp-stage-schedule}.
Each phase is initialized from the preceding phase's checkpoint; step
counts are local to each phase, and learning rates are base values before
scheduling.

\begin{table}[t]
\centering
\small
\setlength{\tabcolsep}{3pt}
\begin{tabular}{lrrr}
\toprule
Phase & Steps & Base LR & Batch \\
\midrule
A & 80,000 & $5\times10^{-5}$ & 2 \\
B & 50,000 & $6.25\times10^{-6}$ & 2 \\
C & 10,000 & $5\times10^{-6}$ & 2 \\
D & 10,000 & $5\times10^{-7}$ & 2 \\
\bottomrule
\end{tabular}
\caption{Stage II training schedule for the model used in the reported
experiments. Steps are counted within each successive phase.}
\label{tab:supp-stage-schedule}
\end{table}

The task is sampled once per training step according to the phase-specific
probabilities in Table~\ref{tab:supp-task-mixture}. Here, clean enhancement
maps $z_{ls}\!\to z_{gs}$, blurred enhancement maps
$z_{lb}\!\to z_{gb}$, bright (well-exposed) deblurring maps $z_{gb}\!\to z_{gs}$,
and joint restoration maps $z_{lb}\!\to z_\alpha$ (to $z_{gs}$ in
Phase D). For Phase C joint samples, the progressive $\alpha$ sampling
described in the main paper is applied after selecting the task.

\begin{table}[t]
\centering
\small
\setlength{\tabcolsep}{4pt}
\begin{tabular}{lp{0.72\columnwidth}}
\toprule
Phase & Task sampling probabilities \\
\midrule
A & 50\% clean enhancement; 50\% blurred enhancement \\
B & 50\% bright deblurring; 25\% clean enhancement; 25\% blurred enhancement \\
C & 60\% joint restoration; 20\% clean enhancement; 10\% blurred enhancement; 10\% bright deblurring \\
D & 70\% joint restoration; 20\% clean enhancement; 10\% bright deblurring \\
\bottomrule
\end{tabular}
\caption{Task sampling during the four MeanFlow training phases. The
percentages in each row sum to 100\%.}
\label{tab:supp-task-mixture}
\end{table}

\paragraph{Inference.}
The observed RAW is encoded once with its domain and illumination condition.
At $(r,t)=(0,1)$, the MeanFlow network is evaluated once with $z_t=z_{\rm in}$.
Its gated velocity is subtracted from the source latent. The result is
decoded with the well-exposed reference condition and rendered by the same
fixed ISP used in the evaluation protocol. No stochastic latent sampling or
iterative denoising is required for this one-step prediction.

%% file: supp/C_baseline_reproduction.tex
\section{Baseline Reproduction and Evaluation Protocol}
\label{sec:supp-baselines}

\paragraph{Data and rendering protocol.}
All methods use the same scene split and per-level test IDs at $0.01$,
$0.001$, and $0.0001$ lux, and the clean and motion-degraded observations
of a scene share the same sharp reference. RAW-input methods receive the
constructed RAW measurements, whereas RGB-input methods are trained and
tested on images rendered from those measurements by the same frozen
MetaISP~\cite{Chen_2024_MetaISP}. The same MetaISP renders every RGB training target and evaluation
reference from the sharp RAW reference. Methods that restore RAW are
rendered by MetaISP before evaluation, while methods that directly predict
RGB (including SIED) are compared with the rendered reference without a
second ISP pass. Each baseline keeps the training objective of its original
method.

\paragraph{Training pairs on SIDED.}
Each baseline category is trained on SIDED for the task it was designed to
solve, and all methods are evaluated against the sharp well-exposed
reference $x_g$. LEAD-MF, in contrast, uses one set of weights for all
benchmarks, illumination levels, and tasks (Sec.~\ref{sec:supp-architecture}).
\begin{itemize}
\item \emph{Enhancement-only methods} (SID, LLPackNet, RRT, DNF, RAWMamba,
LFM-LLIE, and SIED) are trained on the blur-preserving enhancement pairs
$(x_l^{\rm m},x_g^{\rm m})$ of the SIDED training split. They therefore
see the same motion-degraded low-light inputs as all other methods but are
supervised only to restore exposure and appearance and to suppress noise,
not to remove blur.
\item \emph{Joint methods} (DarkIR and LEDNet) explicitly model low-light
blur and are trained on the joint pairs $(x_l^{\rm m},x_g)$, separately at
each illumination level.
\item \emph{Cascades} reuse the SIED and RAWMamba enhancement models above
as their first stage and append EVSSM, which is trained independently on
MetaISP-rendered pairs $(x_g^{\rm m},x_g)$ from the SIDED training split.
\end{itemize}

\paragraph{Rationale.}
The enhancement-only methods contain no mechanism for camera motion; for
them, motion blur is an accompanying degradation of handheld capture rather
than part of the task. Training them on blur-preserving pairs keeps their
supervision consistent with this design: a deblurring target would ask an
architecture without any motion model to solve a task it was never built
for, and would no longer represent the enhancement-only paradigm. These
rows therefore measure what current extremely low-light enhancers deliver
when shake accompanies capture, which is the failure mode illustrated in
Fig.~1(a) of the main paper. Deblurring-aware training is covered by the
other two categories: the joint methods receive full joint supervision, and
the cascades complement each enhancement model with a dedicated deblurring
network. This setting also makes the cascade decomposition consistent:
the first stage maps $x_l^{\rm m}$ toward the blurred well-exposed image
$x_g^{\rm m}$, which matches the input distribution on which EVSSM is
trained. On SIED, which contains no motion blur, all compared methods are
trained on the clean pairs $(x_l,x_g)$.

\paragraph{Optimization budget.}
Each compared model is trained for $200{,}000$ optimizer updates while
retaining its architecture, input domain, loss terms, optimizer, and base
learning rate; the common update budget is the only controlled change
(Table~\ref{tab:supp-baselines}). LFM-LLIE follows the original RGB-domain
latent flow-matching design with a frozen VAE and iterative ODE inference
\cite{Ciubotariu_2026_CVPRW}. For the two-stage SIED model, the two
learning rates correspond to its first and second stages.

\begin{table*}[t]
\centering
\small
\setlength{\tabcolsep}{4pt}
\begin{tabular}{llrrll}
\toprule
Method & Input domain & Updates & Batch / crop & Optimizer and base LR & Training objective \\
\midrule
SID & RAW & 200k & 20 / 512 & Adam, $1\times10^{-4}$ & RGB L1 \\
LLPackNet & RAW & 200k & 128 / 512 & AdamW, $1\times10^{-4}$ & RGB MSE, L1, SSIM, SAM \\
RRT & RAW & 200k & 8 / 512 & Adam, $1\times10^{-4}$ & RGB L1, MS-SSIM \\
DNF & RAW & 200k & 1 / 1024 & AdamW, $2\times10^{-4}$ & RAW and RGB L1 \\
RAWMamba & RAW & 200k & 3 / 512 & AdamW, $1\times10^{-4}$ & RAW and RGB L1 \\
LFM-LLIE & RGB & 200k & 8 / 2048 & AdamW, $1\times10^{-4}$ & Latent flow matching \\
SIED & RAW & 200k & 1 / 512 & Adam, $1\times10^{-4}$ / $8\times10^{-5}$ & Content/contrastive, then diffusion \\
DarkIR & RGB & 200k & 8 / 256 & AdamW, $2\times10^{-4}$ & Charbonnier, SSIM, edge, side \\
LEDNet & RGB & 200k & 8 / 256 & Adam, $1\times10^{-4}$ & Pixel, perceptual, side \\
EVSSM & RGB & 200k & 2 / 128 & AdamW, $1\times10^{-3}$ & Pixel L1, Fourier L1 \\
\bottomrule
\end{tabular}
\caption{Training configurations of the reproduced baselines.
Batch denotes the effective training batch. The update budget is shared,
while method-specific optimization objectives and learning rates are
retained. EVSSM is the independently trained deblurring component of the
SIED $\rightarrow$ EVSSM and RAWMamba $\rightarrow$ EVSSM cascades.}
\label{tab:supp-baselines}
\end{table*}

\paragraph{Cascades, and metrics.}
At inference, each cascade passes the RGB output of its SIDED-trained
enhancement model (SIED outputs RGB directly; RAWMamba is rendered by
MetaISP) to EVSSM without joint fine-tuning, and the EVSSM output is
evaluated directly in RGB space. Appending EVSSM improves SIED by
$0.40$ dB at $0.01$ lux but lowers RAWMamba by $1.23$ dB: residual noise,
color shifts, and over-smoothing from the enhancement stage fall outside
the blur model that EVSSM learns from well-exposed pairs, so errors of
the first stage are not corrected and can be amplified. This illustrates
the limitation of separately trained stages that motivates our unified
model. 

%% file: supp/D_additional_ablations.tex
\section{Additional Ablation Details and Analysis}
\label{sec:supp-ablations}

This section details the protocol of each ablation in the main ablation
table and analyzes what each result implies for the design of LEAD-MF.
Table~\ref{tab:supp-ablation-summary} collects the effect of every
component. Two consistency checks apply throughout. First, the full-model
entries in parts (b)--(d) of the main ablation table ($24.15$ dB /
$0.764$ / $0.060$) coincide with the $0.01$-lux SIDED result in the main
comparison, and the clean-enhancement entry in part (c) ($26.10$ dB)
coincides with the $0.01$-lux SIED result; all ablations therefore use the
final checkpoint and the same test split. Second, each comparison changes
a single factor while keeping the tokenizer, backbone, curriculum, ISP, and
evaluation protocol fixed.

\begin{table}[t]
\centering
\footnotesize
\setlength{\tabcolsep}{2.5pt}
\begin{tabular}{@{}llcccl@{}}
\toprule
Component & Evaluation & $\Delta$PSNR & $\Delta$SSIM & $\Delta$SAM & Overhead \\
\midrule
DIJAM & Tok.\ recon. & $+3.99$ & $+0.023$ & $-0.017$ & Affine/stage \\
$\mathcal L_{ret}$ & SIDED joint & $+1.42$ & $+0.007$ & $-0.006$ & None \\
Two heads & SIED clean & $+2.50$ & $+0.014$ & $-0.002$ & One conv \\
Two heads & SIDED joint & $+2.06$ & $+0.051$ & $-0.011$ & One conv \\
$u_{\rm deb}$ on & SIDED joint & $+1.89$ & $+0.040$ & $-0.001$ & None \\
\bottomrule
\end{tabular}
\caption{Effect of each component, computed from the main ablation table
at $0.01$ lux (except the tokenizer reconstruction). Positive
$\Delta$PSNR/$\Delta$SSIM and negative $\Delta$SAM favor the component.
Overhead refers to inference: DIJAM adds one channel-wise affine
modulation per tokenizer stage, the second head adds one output
convolution on the shared decoder feature, and $\mathcal L_{ret}$ and the
$u_{\rm deb}$ intervention add no inference cost.}
\label{tab:supp-ablation-summary}
\end{table}

\paragraph{MeanFlow versus direct regression.}
The direct-regression variant in Fig.~1(b) of the main paper uses the same
frozen tokenizer, conditional backbone, residual-head composition, four-phase
task curriculum (Table~\ref{tab:supp-task-mixture}), phase lengths, batch
size, and phase-specific learning rates as LEAD-MF
(Table~\ref{tab:supp-stage-schedule}). Only the objective used to learn the
latent displacement in Phases A--C changes: instead of the MeanFlow
average-velocity objective and its Jacobian--vector product, direct
regression minimizes
\begin{equation}
\mathcal L_{\rm reg}
=\left\|u_\theta(z_{\rm in},z_{\rm in},1,1,c_{\mathrm{mf}})
       -(z_{\rm in}-z_{\rm tar})\right\|_2^2.
\end{equation}
In Phase D, both variants use the RGB and Retinex refinement objective of
the main paper, and at inference both subtract one network prediction from
the source latent and decode with the same tokenizer and ISP. The two
variants therefore have identical architecture and inference cost, and any
difference reflects the training objective alone. Direct regression
supervises only the endpoint $(r,t)=(0,1)$, whereas MeanFlow samples
$(r,t)$ along the whole source--target path and enforces the
average-velocity identity through the Jacobian--vector product, which
constrains the learned displacement field beyond its endpoint while still
permitting one-step inference. The flow-matching (LFM) and DDPM (SIED)
entries in the same panel correspond to the LFM-LLIE and SIED baselines of
the main paper.

\paragraph{Tokenizer with and without DIJAM.}
\emph{Protocol.} Part (a) of the main ablation table evaluates standalone
RAW self-reconstruction before the restoration network is trained. Both
variants use the same dataset-adaptive low-light scaling, shared encoder and
decoder, latent dimensions, low-light and well-exposed RAW inputs, and
reconstruction objective. The unconditioned variant disables domain and
illumination modulation throughout the encoder and decoder, whereas the
DIJAM variant applies the stage-wise modulation described in
Sec.~\ref{sec:supp-architecture}. Each input is reconstructed in its own
RAW domain, and neither variant uses MeanFlow or a restoration target.

\emph{Analysis.} DIJAM improves reconstruction by $3.99$ dB PSNR and
$0.023$ SSIM, and reduces SAM by $22.7\%$ ($0.075\rightarrow0.058$).
This pattern is consistent with the form of DIJAM, whose channel-wise
scale and bias can compensate for the domain- and illumination-dependent
channel statistics that a single unconditioned encoder--decoder must
otherwise absorb. The result matters for
restoration because every LEAD-MF prediction is decoded by the same
tokenizer: with DIJAM, reconstruction reaches $50.23$ dB, far above the
$24$--$26$ dB restoration accuracy at $0.01$ lux, so decoding error is not
the limiting factor and the compact latent space can be used for
full-image inference without sacrificing fidelity. DIJAM adds only one
affine modulation per stage and no separate encoder or decoder per
illumination level.

\paragraph{Retinex consistency in Phase D.}
\emph{Protocol.} Both variants in part (b) start from the same Phase-C
restoration model and apply the Phase-D refinement with the same task
curriculum, architecture, and frozen tokenizer and ISP. The full variant
retains the frozen decomposition network and the term
$\lambda_{\rm ret}\mathcal L_{\rm ret}$; the control sets
$\lambda_{\rm ret}=0$ while keeping the RGB MSE and SAM terms. Both are
evaluated on joint enhancement--deblurring of the same $0.01$-lux SIDED test
scenes, and the full variant is the final model reported in the main paper.

\emph{Analysis.} Retinex consistency improves PSNR by $1.42$ dB and reduces
SAM by $9.1\%$, while SSIM changes by only $0.007$. A large PSNR and SAM gain
with a small SSIM change indicates that the term mainly corrects global
intensity and chromaticity rather than local structure, which matches its
construction: the illumination term constrains the smooth exposure map and
the reflectance term constrains color and albedo, complementing the
pixel-wise MSE and the angle-based SAM loss. Because the decomposition
network is used only as a frozen training-time critic, this gain incurs no
inference overhead.

\begin{figure*}[p]
    \centering
    \includegraphics[width=0.90\textwidth]{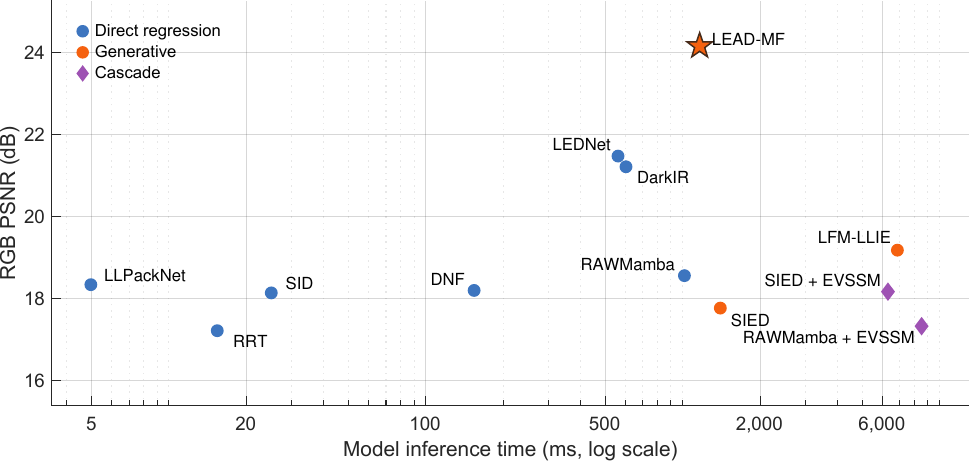}
    \caption{RGB PSNR versus mean model-only inference time on $0.01$-lux
    SIDED. All methods process the same image. Timings cover the end-to-end
    model pipeline on an RTX 4090, including tokenizer encoding and decoding
    for LEAD-MF, and exclude ISP for every method; the horizontal axis uses
    a logarithmic scale. Blue points are
    direct-regression methods, orange points are generative methods, and
    purple diamonds are cascaded methods. The orange star highlights
    LEAD-MF. Each method uses its own input resolution and batch
    configuration.}
    \label{fig:supp-latency-psnr}
\end{figure*}

\begin{figure*}[p]
    \centering
    \includegraphics[page=1,width=0.50\textwidth]{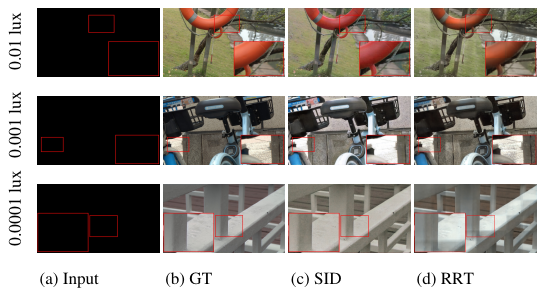}
    \vspace{2mm}
    \includegraphics[page=2,width=0.94\textwidth]{Figures/additional_qualitative_results.pdf}
    \caption{Additional same-scene qualitative comparisons omitted from the
    main figures. Top: SIED without motion blur. Bottom: SIDED with motion
    blur. The rows correspond to $0.01$, $0.001$, and $0.0001$ lux; Input
    and GT are repeated for reference. All method outputs use the same
    red-box regions and enlarged views as the main figures.}
    \label{fig:supp-additional-qualitative}
\end{figure*}

\paragraph{Single-head architectural ablation.}
\emph{Protocol.} For part (c), we retrain a control model whose shared
backbone terminates in one output head that predicts the complete
task-conditioned latent velocity, whereas LEAD-MF combines two parallel
outputs as $u_{\theta}=u_{\rm enh}+b\alpha u_{\rm deb}$. The control
retains the same tokenizer, backbone, task conditioning, four-phase
curriculum, losses, and one-step inference; only the output-head design
changes. Both models are evaluated with identical conditions on two tasks
at $0.01$ lux: clean enhancement on SIED ($b=0$), where the deblurring
contribution of LEAD-MF is gated off, and joint restoration on the 180
SIDED test scenes ($b=\alpha=1$), where both heads are active. The
comparison assesses two independently trained architectures under a matched
curriculum and complements the inference-time intervention in part (d).

\emph{Analysis.} The single-head model obtains $23.60$ dB / $0.801$ /
$0.060$ (PSNR / SSIM / SAM) on clean enhancement and $22.09$ dB / $0.713$ /
$0.071$ on joint restoration, versus $26.10$ dB / $0.815$ / $0.058$ and
$24.15$ dB / $0.764$ / $0.060$ for LEAD-MF. Three observations support the
decomposition. (i)~LEAD-MF is better on \emph{both} tasks, by $2.50$ and
$2.06$ dB PSNR. Since the gain persists when the deblurring head is gated
off, it cannot be explained by the extra head alone; it reflects a better
use of the shared backbone. (ii)~The SSIM and SAM gaps are larger on joint
restoration ($+0.051$ and $-0.011$) than on clean enhancement ($+0.014$
and $-0.002$), i.e., the advantage grows when structural deblurring is
required. (iii)~The single-head joint result is even below the head-off
output of LEAD-MF ($22.09$ vs.\ $22.26$ dB) in part (d), so a single output
that must encode both residuals is worse than the enhancement residual of
LEAD-MF alone, indicating interference between the two tasks.

The design explains why the decomposition helps. With gating
$b\alpha$, the curriculum gives each head targeted supervision: in Phase A,
blur-preserving enhancement pairs $(z_{lb},z_{gb})$ use $\alpha=0$, so the
gradient reaches only $u_{\rm enh}$, which learns exposure and appearance
recovery without being asked to remove blur; in Phase B, well-exposed
deblurring pairs $(z_{gb},z_{gs})$ activate $u_{\rm deb}$ on inputs whose
structural cues have high SNR. The two heads thus mirror the latent path
decomposition $z_{lb}-z_\alpha=(z_{lb}-z_{gb})+(z_{gb}-z_\alpha)$ of the main
paper. A single head receives the same conditions but must infer from them
when to include the deblurring component, so the two residuals compete for
the same output. The decomposition adds only one output convolution and
keeps a single backbone evaluation.

\paragraph{Deblurring-output intervention.}
\emph{Protocol.} For part (d), we evaluate the final two-head checkpoint on
the 180 $0.01$-lux SIDED test scenes with the joint condition
$c_{\mathrm{mf}}^{\mathrm{joint}}$ ($b=\alpha=1$) fixed and compute both
heads in the ordinary one-step forward pass:
\begin{equation}
\begin{aligned}
(u_{\rm enh},u_{\rm deb})
  &=F_\theta(z_{\rm in},z_{\rm in},1,1,c_{\mathrm{mf}}^{\mathrm{joint}}),\\
\hat z_g^{\rm full}&=z_{\rm in}-u_{\rm enh}-u_{\rm deb},\\
\hat z_g^{\rm off}&=z_{\rm in}-u_{\rm enh}.
\end{aligned}
\end{equation}
The intervention changes neither the checkpoint nor the network condition,
and both predictions are decoded with the same frozen GT-domain decoder and
ISP.

\emph{Analysis.} Against the sharp reference $x_g$, the head-off output
obtains $22.26$ dB PSNR, $0.724$ SSIM, and $0.061$ SAM, whereas the full
output obtains $24.15$ dB, $0.764$, and $0.060$. Against the blurred
normal-exposure reference $x_g^{\rm m}$, the head-off output obtains
$27.84$ dB, $0.928$, and $0.047$. These numbers reveal the division of
labor between the heads. First, the head-off output is $5.58$ dB closer to
$x_g^{\rm m}$ than to $x_g$, so $u_{\rm enh}$ alone produces an
exposure-restored but still blurred image, as supervised in Phase A.
Second, adding $u_{\rm deb}$ raises sharp-reference PSNR by $1.89$ dB and
SSIM by $0.040$ while changing SAM by only $0.001$; the correction therefore
acts mainly on spatial structure rather than on color or exposure, which is
the expected signature of motion deblurring. Together with part (c), these
results show that the deblurring head carries the motion-restoration
correction and that separating it from the enhancement residual benefits
both tasks.

\paragraph{Summary.}
Each component addresses a distinct failure mode at little or no inference
cost: DIJAM aligns the latent space across exposure domains and darkness
levels, Retinex consistency corrects global
intensity and chromaticity, and the
dual-head decomposition improves structure under motion blur while also benefiting clean enhancement.

%% file: supp/E_efficiency.tex
\section{Inference Efficiency}
\label{sec:supp-efficiency}

SIED and LFM-LLIE are the two highest-PSNR baselines for clean enhancement
at $0.01$ lux. On the corresponding motion-degraded SIDED test set,
one-step LEAD-MF achieves $24.15$ dB in $1164$ ms per image
(Fig.~\ref{fig:supp-latency-psnr}). It gains $6.38$ dB over SIED
($17.77$ dB, $1400.90$ ms) and $4.97$ dB over LFM-LLIE
($19.18$ dB, $6848$ ms), while reducing their inference times by $16.9\%$
and $83.0\%$, respectively.

LEAD-MF also improves PSNR by $5.98$ dB over SIED $\rightarrow$ EVSSM and
$6.82$ dB over RAWMamba $\rightarrow$ EVSSM, with respective inference-time
reductions of $81.5\%$ and $86.3\%$. Every method with a shorter
inference time trails LEAD-MF by at least $2.68$ dB PSNR (LEDNet), and
LEAD-MF is the only method above $22$ dB, lying on the quality--efficiency
Pareto front. Timings are mean model-only inference times per image on an
NVIDIA GeForce RTX 4090 with each method's own input resolution and batch
configuration, excluding ISP.

%% file: supp/F_additional_qualitative.tex
\section{Additional Qualitative Comparisons}
\label{sec:supp-additional-qualitative}

Figure~\ref{fig:supp-additional-qualitative} adds the baselines absent
from the main qualitative figures (SIED: SID and RRT; SIDED: SID, DNF,
LLPackNet, SIED, RRT, and RAWMamba $\rightarrow$ EVSSM), using the same
scenes, illumination levels, and enlarged regions.